\pdfoutput=1

\documentclass[11pt]{article}

\usepackage{ACL2023}

\usepackage{times}
\usepackage{latexsym}
\usepackage{booktabs}
\usepackage{graphicx}
\usepackage{hyperref}

\usepackage{tikz}
\def\checkmark{\tikz\fill[scale=0.4](0,.35) -- (.25,0) -- (1,.7) -- (.25,.15) -- cycle;} 

\usepackage[T1]{fontenc}

\usepackage[utf8]{inputenc}

\usepackage{microtype}

\usepackage{inconsolata}

\newcommand\blfootnote[1]{%
    \begingroup%
\let\thefootnote\relax\footnotetext{\hspace{-4pt}#1}%
\endgroup}%

\title{Epicurus at SemEval-2023 Task 4: Improving Prediction of
Human Values behind Arguments by Leveraging Their Definitions}

\author{Christian Fang $\star$ \\
  Department of Sociology \\
  Utrecht University \\
  \texttt{c.fang@uu.nl} \\\And
  Qixiang Fang $\star$\\
  Department of Methodology \\ and Statistics \\ 
  Utrecht University \\
  \texttt{q.fang@uu.nl} \\ \And
  Dong Nguyen \\
  Department of Information \\ and Computing Sciences \\
  Utrecht University \\
  \texttt{d.p.nguyen@uu.nl}}

\begin{document}
\maketitle

\begin{abstract}
We describe our experiments for SemEval-2023 Task 4 on the identification of human values behind arguments (ValueEval).
Because human values are subjective concepts which require precise definitions, we hypothesize that incorporating the definitions of human values (in the form of annotation instructions and validated survey items) during model training can yield better prediction performance. We explore this idea and show that our proposed models perform better than the challenge organizers' baselines, with improvements in macro $F_1$ scores of up to 18\%. 
\end{abstract}

\blfootnote{$\star$ Shared first authorship.}

\section{Introduction}
Human values are distinct beliefs that guide human behavior \cite{schwartz2012refining}. Examples of such values are hedonism (i.e., seeking pleasure in life), face (i.e., maintaining recognition in society), and humility (i.e., being humble). 

Studying human values has a long history in the social sciences and in studies of formal argumentation due to their manifold applications \cite{kiesel:2022, schwartz2012refining}. For example, researchers might be interested in studying how the human values individuals subscribe to affect their charitable behavior \cite{sneddon2020personal} or voting behavior \cite{barnea1998values}. In NLP, human values can be leveraged for personality recognition \cite{maheshwari2017societal}, or for assessing what values are implied in online discourses \cite{kiesel:2022}. To that end, Task 4 (ValueEval) of the SemEval 2023 competition \cite{kiesel:2023} called for participants to design NLP systems that can classify a given argument as belonging to one of 20 value categories described in Kiesel et al.'s human value taxonomy \citeyearpar{kiesel:2022}. 

Human values are inherently subjective concepts, which becomes evident in, for example, the existence of many human value taxonomies, each of which contains somewhat different and differently-defined human values \citep[e.g.][]{rokeach1973nature, schwartz2012refining}. Accordingly, in any scientific study of human values, clear definitions are key. Therefore, we argue that it is important to incorporate the definitions of human values into model training. Our approach leverages the definitions of human values (based on survey questions and annotation instructions), which we refer to as \textbf{definitional statements}, in a natural language inference (NLI) setup. This approach offers additional theoretical benefits, such as potentially higher model validity (i.e., more accurate encodings of human values) as well as greater prediction reliability (see~\S\ref{sec:measurement}). We showed that our approach achieved better performance than the challenge baselines. We also conducted additional post-hoc analyses to explore how prediction performance would vary by the number of definitional statements per value category. We found that even with only a few definitional statements per value category, our proposed approach achieved good performance. Our code is available in a public GitHub repository (\url{https://github.com/fqixiang/SemEval23Task4}).

\section{Background}
\subsection{Task Setup}
\begin{figure}[h]
\centering
\includegraphics[width=75mm,scale=0.7]{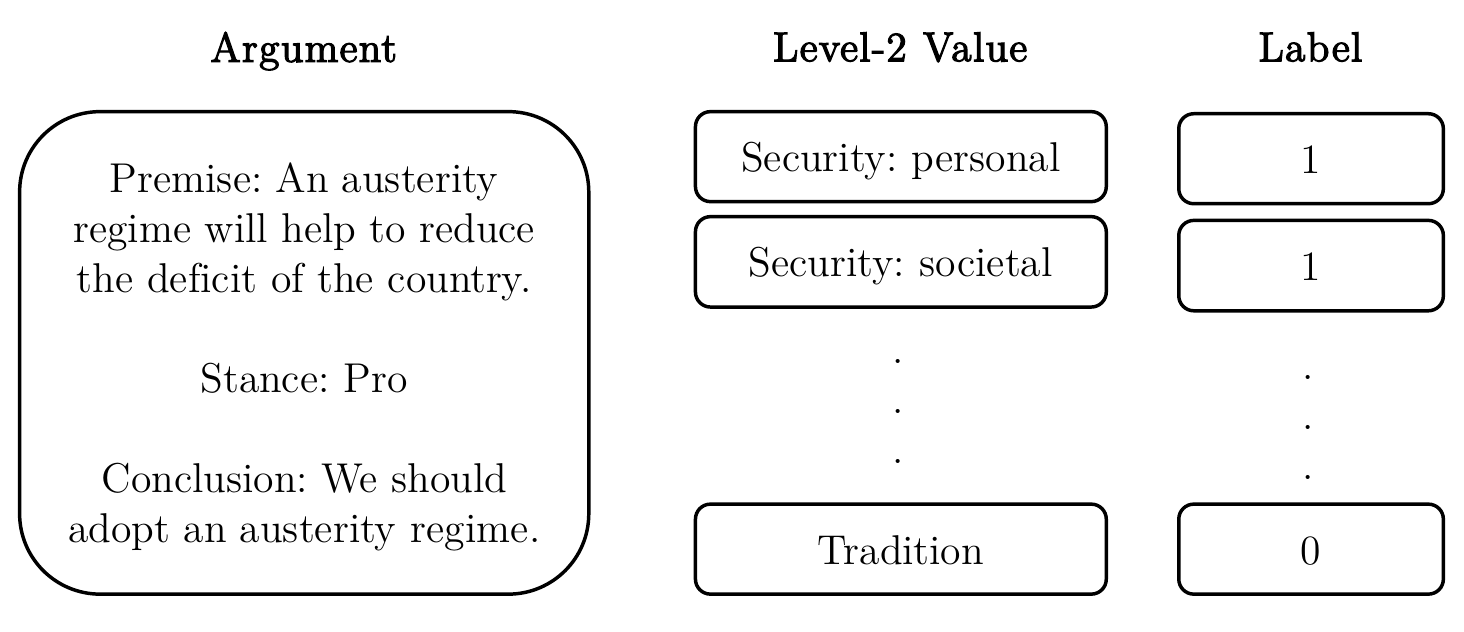}
\caption{Illustration of the task.}
\label{figure1-data}
\end{figure}

The goal of the challenge task was to, given a textual argument and a human value category, classify whether the argument entails that value category. Each argument consisted of a premise, stance (for or against), and conclusion and was assigned a binary label for each of the 20 human value categories (also called level-2 values in \citet{kiesel:2022}). Figure~\ref{figure1-data} illustrates this. 

\subsection{Related Work}
Our approach of using definitions of human values in the form of annotation instructions and survey questions is positioned within two streams of prior literature.
The first stream used dictionary definitions and annotation instructions to improve, for instance, information extraction from legal documents (especially with a small data set) \cite{kang2021label}, retrieval of the semantic roles of tokens \cite{zhang2022label}, and detection of slang words \cite{wilson2020urban}, dialects~\cite{demszky2020learning}, and rare words~\cite{pilehvar2017inducing}. 
The second stream used survey questions for improving the prediction of social science constructs, such as personality traits \cite{kreuter2022items, vu2020predicting, yang2021learning}, as well as social and political attitudes \cite{fang2022modelling}. 

\section{A Measurement Problem}
\label{sec:measurement}
In the social sciences, detecting human values is a measurement problem and typically relies on the use of (validated) survey instruments, such as the Schwartz Value survey~\cite{schwartz2012refining}. Typically, respondents are asked to rate the importance of a given value (e.g., hedonism), presented in the form of survey questions, on a numerical scale. In this way, a numerical summary per human value can be assigned to each respondent.

Because human values are abstract concepts that are not directly observable, the respective measurements are likely to suffer from measurement error \cite{Fang2022EvaluatingTC}. Social scientists are, therefore, particularly concerned about the content validity and reliability of such measurements. Content validity refers to an instrument fully capturing what it aims to measure \cite{trochim2016research}, whereas reliability means that measurements are stable over time and contexts (i.e., do not suffer from large random variations) \cite{trochim2016research}. To ensure high content validity, multiple survey questions that capture different sub-aspects of a value (e.g., hedonism) are typically used. Each respondent's answers to these questions about the same value are then aggregated (e.g., averaged) to obtain a single, more reliable score.

Likewise, incorporating definitional statements when training a model to predict human values from arguments might help to improve the content validity of the model, as well as the reliability of the predictions~\cite{fang2022modelling}. For instance, the human value "achievement" has many sub-aspects (i.e., being ambitions, having success, being capable, being intellectual, and being courageous). By incorporating these finer-grained definitions of "achievement" into model training, the model can learn to encode the full scope of this value, which can in turn help to identify arguments that entail this value. Furthermore, averaging a model’s predictions across multiple definitional statements of the same human value category can lead to more reliable, less random results, which is consistent with the social sciences' approach. 

\section{System Overview}
\subsection{Data Augmentation with Definitional Statements}
We created definitional statements for each of the 20 value categories based on two sources. The first source were the annotation instructions, which we obtained from the annotation interface provided by \citet{kiesel:2022}. 
The second source were the survey questions that underlie the human value taxonomy uses in this challenge. We collected all relevant survey questions from the surveys that Kiesel et al. \citeyearpar{kiesel:2022} based their human value taxonomy on, namely the PVQ5X Value Survey \cite{schwartz2012refining}; its predecessor, the Schwartz Value survey \cite{schwartz1992universals}; the World Value Survey \cite{haerpfer2022world}; the Rokeach Value Survey \cite{rokeach1973nature}; and the Life Values Inventory \cite{brown2002life}.
For an overview of the number of definitional statements per value category, see Appendix~\ref{append_b}. 

\begin{figure}[h!]
\centering
\includegraphics[width=75mm,scale=0.7]{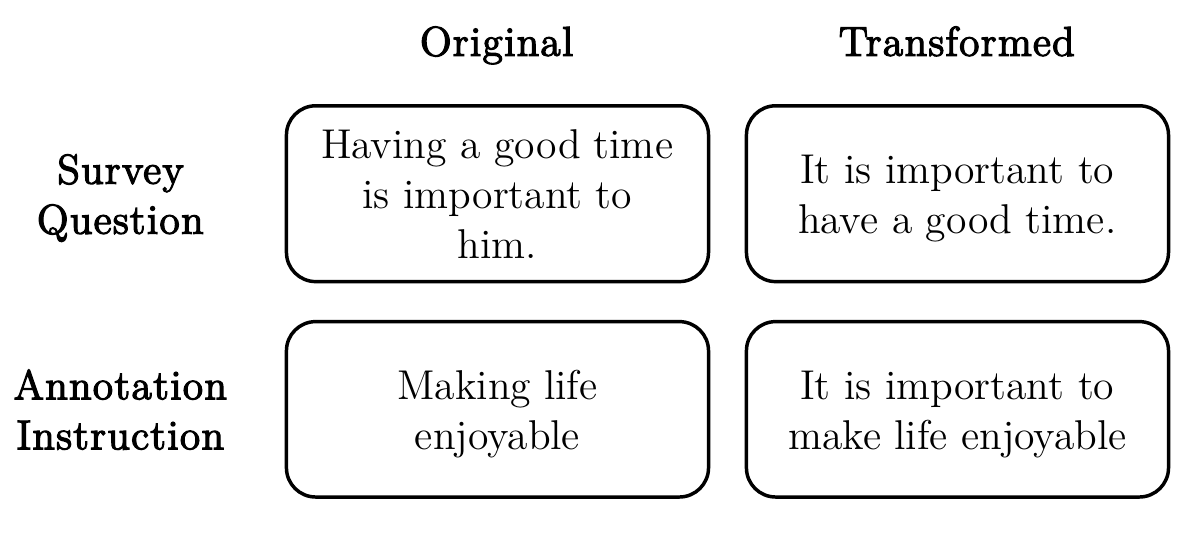}
\caption{Example for original and transformed definitional statements for the value category "hedonism". }
\label{figure2-ds}
\end{figure}

We harmonized all definitional statements by forcing them to adhere to a "It is important to be/have" sentence structure to prevent models from learning uninformative idiosyncratic formulations. Figure~\ref{figure2-ds} shows such as an example.

Next, we augmented the training data set with definitional statements. We dropped "conclusion" and "stance" from the arguments, because per our observation of the  data, the "premise"s alone already contain all the information about the underlying human values, which renders the use of "conclusion" and "stance" redundant. Each premise in the data set was combined with each definitional statement. For each combination of premise and definitional statement, we assigned "entailment" if the associated value label in the training data was 1, and "not entailment" otherwise.

\subsection{NLI Setup}
We used an NLI setup for modelling. NLI involves judging whether a hypothesis can be inferred from a premise. If so, then that premise entails the hypothesis. In our case, the textual premises from arguments constitute the premises, whereas the definitional statements constitute the hypotheses. We used BERT \cite{devlin2018bert} as the model of choice within the NLI setup.

\subsection{Averaging Predictions and Thresholding}
\begin{figure}[h]
\centering
\includegraphics[width=75mm,scale=0.7]{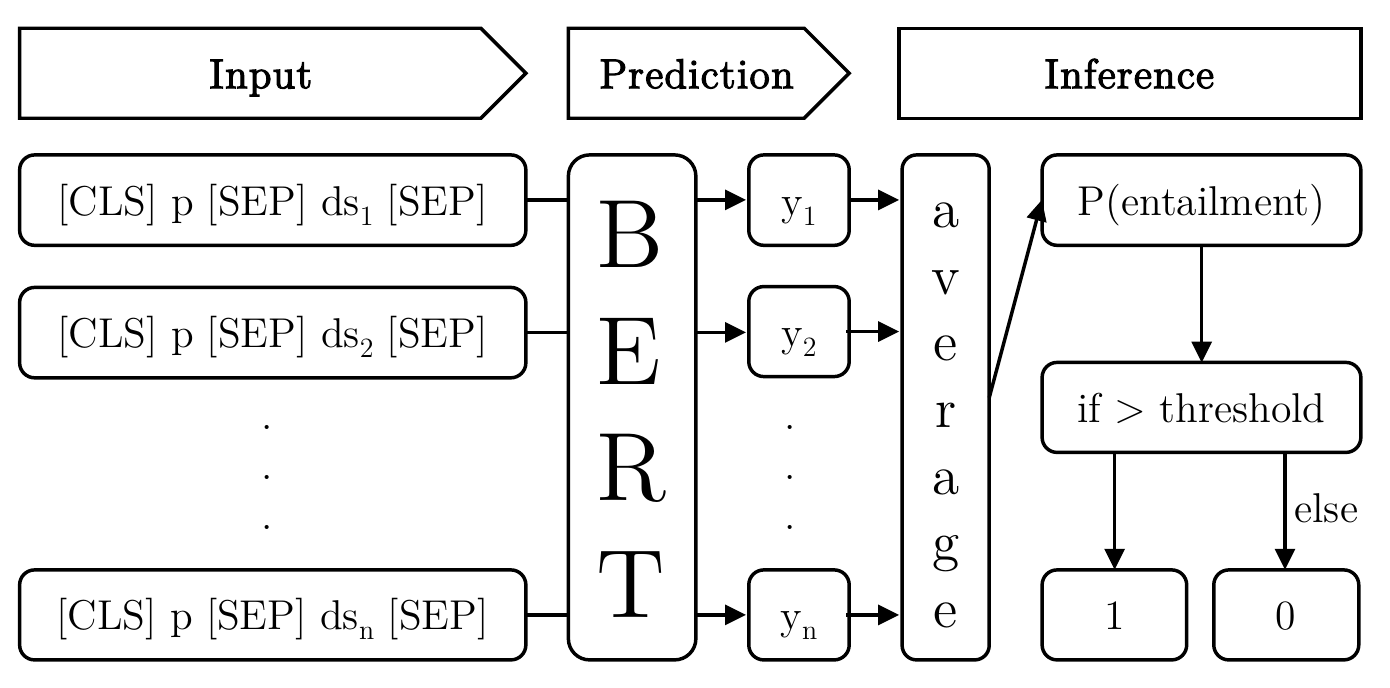}
\caption{Illustration of the prediction step of the system for a given premise ($p$) and a value category described by $n$ definitional statements ($ds_i$). We feed each combination of $p$ and $ds_i$ into a finetuned BERT model, obtain each individual binary prediction $y_i$, average these predictions to obtain the probability of entailment for a value category given the premise, and finally, make a binary decision based on a (finetuned) threshold.}
\label{figure3_inference}
\end{figure}

Our system yields a binary prediction for each combination of premise and definitional statement. Therefore, multiple predictions exist for every combination of premise and human value. To convert these multiple predictions into a single binary prediction per premise and value category, we averaged the predictions per value category, and applied a (fine-tuned) threshold to determine whether it is an entailment. Figure~\ref{figure3_inference} illustrates this.


\section{Experimental Setup}
\subsection{Data Splits}
The main data set comprised 8,865 annotated arguments. We used the same split as the challenge organizers, namely a training set (61\%), a validation set (21\%), and a test set (18\%). The label distribution of this data set was highly imbalanced: for example, only about 3.4\% of all arguments were labelled "hedonism", whereas 47.6\%  were labelled "Universalism: concern". The second test set (Nahj al-Balagha) contained 279 annotated arguments from Islamic religious texts. The label distribution for that test set was slightly less imbalanced than the main data set \citep{mirzakhmedova:2023}.

\subsection{Preprocessing and Hyperparameter Tuning}
We used the pretrained "bert-base-uncased" and its tokenizer from Hugging Face \cite{wolf-etal-2020-transformers}. 
To construct the input vector, we follow Jiang and de Marneffe \citeyearpar{jiang2019evaluating}, where each premise and definitional statement is separated by the [SEP] token. We used a binary classification head to predict whether a given premise entailed a definitional statement, trained the model based on cross-entropy loss, and chose the best model with the lowest loss on the validation set. 

We fine-tuned the number of training steps, with early stopping where the patience parameter was set to 10. 
We also tested ten different thresholds (ranging from 0 to 0.9 with increments of .1) on the validation set and chose the best-performing threshold for the test set based on macro $F_1$ scores on the validation set. The optimal threshold was 0 or 0.3, depending on the model. The full list of hyperparameters is in Appendix~\ref{append_a}.

\subsection{Trained Models}
We fine-tuned four BERT models (see Table~\ref{table-modeloverview}) based on the type of definitional statements (annotation instructions or survey questions) and loss functions (weighted or unweighted). Weighted cross-entropy loss is considered to account for the imbalanced class distribution in the training data. The weights were calculated as proportional to inverse class distributions in a training batch.

\begin{table}[]
\resizebox{\columnwidth}{!}{%
\begin{tabular}{p{1.7cm}|p{2cm}p{1.5cm}p{1.5cm}}
\textbf{Model} & \textbf{Definitional statements} & \textbf{Weighted loss?} & \textbf{Training size} \\ 
\toprule
\textbf{ANN\textsubscript{uw}} & annotations  &                           & 31,807,742 \\
\textbf{ANN\textsubscript{w}} & annotations & \checkmark &  31,807,742\\
\textbf{SVY\textsubscript{uw}} & survey items        &                           & 37,109,033 \\
\textbf{SVY\textsubscript{w}} & survey items        & \checkmark & 37,109,033
\end{tabular}%
}
\caption{Overview of trained models}
\label{table-modeloverview}
\end{table}


\subsection{Evaluation Metrics}
We evaluated model performance by the macro $F_1$ score, calculated as the average of all 20 individual $F_1$ scores. 
During model training and validation, we were not aware that the challenge organizers used a different method for calculating macro $F_1$, namely by using the averages of precision and recall. Therefore, to stay consistent with our training and validation strategy, we focus on discussing the results based on our macro $F_1$ calculation. This calculation method has also been shown to be the more appropriate metric between the two, especially for model rankings and when data is imbalanced \cite{opitz2019macro}. However, for completeness, we show both results and discuss their differences.

\section{Results}

Table 1 shows the results of our four model runs on the main testing data set (Main), as well as the testing set comprising arguments from religious texts (Nahj al-Balagha). 

\begin{table*}[!ht]
\centering\small%
\setlength{\tabcolsep}{2.5pt}%
\begin{tabular}{@{}ll@{\hspace{10pt}}c@{\hspace{5pt}}cccccccccccccccccccccc@{}}
\toprule
\multicolumn{2}{@{}l}{\bf Test set / Approach} & \rotatebox{90}{\bf macro $F_1$ (our own)} & \rotatebox{90}{\bf macro $F_1$ (official)}& \rotatebox{90}{\bf Self-direction: thought} & \rotatebox{90}{\bf Self-direction: action} & \rotatebox{90}{\bf Stimulation} & \rotatebox{90}{\bf Hedonism} & \rotatebox{90}{\bf Achievement} & \rotatebox{90}{\bf Power: dominance} & \rotatebox{90}{\bf Power: resources} & \rotatebox{90}{\bf Face} & \rotatebox{90}{\bf Security: personal} & \rotatebox{90}{\bf Security: societal} & \rotatebox{90}{\bf Tradition} & \rotatebox{90}{\bf Conformity: rules} & \rotatebox{90}{\bf Conformity: interpersonal} & \rotatebox{90}{\bf Humility} & \rotatebox{90}{\bf Benevolence: caring} & \rotatebox{90}{\bf Benevolence: dependability} & \rotatebox{90}{\bf Universalism: concern} & \rotatebox{90}{\bf Universalism: nature} & \rotatebox{90}{\bf Universalism: tolerance} & \rotatebox{90}{\bf Universalism: objectivity} \\
\midrule
\multicolumn{2}{@{}l}{\emph{Main}} \\
& \textcolor{gray}{Best per category} & \textcolor{gray}{.588} & \textcolor{gray}{.59} & \textcolor{gray}{.61} & \textcolor{gray}{.71} & \textcolor{gray}{.39} & \textcolor{gray}{.39} & \textcolor{gray}{.66} & \textcolor{gray}{.50} & \textcolor{gray}{.57} & \textcolor{gray}{.39} & \textcolor{gray}{.80} & \textcolor{gray}{.68} & \textcolor{gray}{.65} & \textcolor{gray}{.61} & \textcolor{gray}{.69} & \textcolor{gray}{.39} & \textcolor{gray}{.60} & \textcolor{gray}{.43} & \textcolor{gray}{.78} & \textcolor{gray}{.87} & \textcolor{gray}{.46} & \textcolor{gray}{.58} \\

& \textcolor{gray}{Best approach} & \textcolor{gray}{.551} &\textcolor{gray}{.56} & \textcolor{gray}{.57} & \textcolor{gray}{.71} & \textcolor{gray}{.32} & \textcolor{gray}{.25} & \textcolor{gray}{.66} & \textcolor{gray}{.47} & \textcolor{gray}{.53} & \textcolor{gray}{.38} & \textcolor{gray}{.76} & \textcolor{gray}{.64} & \textcolor{gray}{.63} & \textcolor{gray}{.60} & \textcolor{gray}{.65} & \textcolor{gray}{.32} & \textcolor{gray}{.57} & \textcolor{gray}{.43} & \textcolor{gray}{.73} & \textcolor{gray}{.82} & \textcolor{gray}{.46} & \textcolor{gray}{.52} \\

& \textcolor{gray}{BERT} & \textcolor{gray}{.391} &\textcolor{gray}{.42}& \textcolor{gray}{.44} & \textcolor{gray}{.55} & \textcolor{gray}{.05} & \textcolor{gray}{.20} & \textcolor{gray}{.56} & \textcolor{gray}{.29} & \textcolor{gray}{.44} & \textcolor{gray}{.13} & \textcolor{gray}{.74} & \textcolor{gray}{.59} & \textcolor{gray}{.43} & \textcolor{gray}{.47} & \textcolor{gray}{.23} & \textcolor{gray}{.07} & \textcolor{gray}{.46} & \textcolor{gray}{.14} & \textcolor{gray}{.67} & \textcolor{gray}{.71} & \textcolor{gray}{.32} & \textcolor{gray}{.33} \\

& \textcolor{gray}{1-Baseline} & \textcolor{gray}{.249} & \textcolor{gray}{.26} & \textcolor{gray}{.17} & \textcolor{gray}{.40} & \textcolor{gray}{.09} & \textcolor{gray}{.03} & \textcolor{gray}{.41} & \textcolor{gray}{.13} & \textcolor{gray}{.12} & \textcolor{gray}{.12} & \textcolor{gray}{.51} & \textcolor{gray}{.40} & \textcolor{gray}{.19} & \textcolor{gray}{.31} & \textcolor{gray}{.07} & \textcolor{gray}{.09} & \textcolor{gray}{.35} & \textcolor{gray}{.19} & \textcolor{gray}{.54} & \textcolor{gray}{.17} & \textcolor{gray}{.22} & \textcolor{gray}{.46} \\

& ANN\textsubscript{uw}* & .431 & .44 & .47 &	.59 & .13 & .15 & .57 &	.33 & .50 &	.29 & .70 &	.59 & .47 &	.54 & .19 &	.21 & .50 &	.19 & .69 &	.72 & .33 &	.45 \\
& ANN\textsubscript{w} & .450 & .46 & .49 & .59 & .22 & .33 & .57 & .36 & .50 & .23 & .70 & .61 & .47 & .45 & .26 & .19 & .47 & .28 & .68 & .74 & .34 & .52 \\
& SVY\textsubscript{uw}* & .434 & .45 &	.46 & .56 & .18 & .31 & .58 & .35 & .55 & .20 & .70 & .58 & .44 & .50 & .11 &.21 &	.48	& .26 & .69 &	.75 & .34 & .45 \\
& SVY\textsubscript{w} & .435 & .44 & .47 & .58 & .21 & .22 & .56 & .32 & .48 & .26 & .70 & .59 & .41 & .47 & .22 & .14 & .48 & .27 & .69 & .72 & .37 & .53 \\

\addlinespace
\multicolumn{2}{@{}l}{\emph{Nahj al-Balagha}} \\
& \textcolor{gray}{Best per category} & \textcolor{gray}{.428} & \textcolor{gray}{.48} & \textcolor{gray}{.18} & \textcolor{gray}{.49} & \textcolor{gray}{.50} & \textcolor{gray}{.67} & \textcolor{gray}{.66} & \textcolor{gray}{.29} & \textcolor{gray}{.33} & \textcolor{gray}{.62} & \textcolor{gray}{.51} & \textcolor{gray}{.37} & \textcolor{gray}{.55} & \textcolor{gray}{.36} & \textcolor{gray}{.27} & \textcolor{gray}{.33} & \textcolor{gray}{.41} & \textcolor{gray}{.38} & \textcolor{gray}{.33} & \textcolor{gray}{.67} & \textcolor{gray}{.20} & \textcolor{gray}{.44} \\

& \textcolor{gray}{Best approach} & \textcolor{gray}{.40} & \textcolor{gray}{.356} & \textcolor{gray}{.13} & \textcolor{gray}{.49} & \textcolor{gray}{.40} & \textcolor{gray}{.50} & \textcolor{gray}{.65} & \textcolor{gray}{.25} & \textcolor{gray}{.00} & \textcolor{gray}{.58} & \textcolor{gray}{.50} & \textcolor{gray}{.30} & \textcolor{gray}{.51} & \textcolor{gray}{.28} & \textcolor{gray}{.24} & \textcolor{gray}{.29} & \textcolor{gray}{.33} & \textcolor{gray}{.38} & \textcolor{gray}{.26} & \textcolor{gray}{.67} & \textcolor{gray}{.00} & \textcolor{gray}{.36} \\

& \textcolor{gray}{BERT} & \textcolor{gray}{.2155} & \textcolor{gray}{.28}& \textcolor{gray}{.14} & \textcolor{gray}{.09} & \textcolor{gray}{.00} & \textcolor{gray}{.67} & \textcolor{gray}{.41} & \textcolor{gray}{.00} & \textcolor{gray}{.00} & \textcolor{gray}{.28} & \textcolor{gray}{.28} & \textcolor{gray}{.23} & \textcolor{gray}{.38} & \textcolor{gray}{.18} & \textcolor{gray}{.15} & \textcolor{gray}{.17} & \textcolor{gray}{.35} & \textcolor{gray}{.22} & \textcolor{gray}{.21} & \textcolor{gray}{.00} & \textcolor{gray}{.20} & \textcolor{gray}{.35} \\

& \textcolor{gray}{1-Baseline} & \textcolor{gray}{.121} & \textcolor{gray}{.13} & \textcolor{gray}{.04} & \textcolor{gray}{.09} & \textcolor{gray}{.01} & \textcolor{gray}{.03} & \textcolor{gray}{.41} & \textcolor{gray}{.04} & \textcolor{gray}{.03} & \textcolor{gray}{.23} & \textcolor{gray}{.38} & \textcolor{gray}{.06} & \textcolor{gray}{.18} & \textcolor{gray}{.13} & \textcolor{gray}{.06} & \textcolor{gray}{.13} & \textcolor{gray}{.17} & \textcolor{gray}{.12} & \textcolor{gray}{.12} & \textcolor{gray}{.01} & \textcolor{gray}{.04} & \textcolor{gray}{.14} \\

& ANN\textsubscript{uw}* & .252 &	.28 &	.10 &	.22 &	.00 &	  .18	 & .52	& .11	& .00	& .55	 & .40 &	.25 & 	.54 & 	.26 &	.24 &	.24 & 	.30 & 	.29 & 	.25	 & .25	& .05	& .28 \\
& ANN\textsubscript{w}& .2155 & .24& .16 & .17 & .00 & .18 & .47 & .08 & .12 & .46 & .37 & .31 & .39 & .15 & .06 & .15 & .31 & .23 & .19 & .13 & .06 & .32 \\
& SVY\textsubscript{uw}* & .254	& .28 &	.10 & .24 &	.00 &  .40	& .50	& .09	& .25	& .52 & .41 & .24	& .44	& .19 &	.10	& .25	& .27	& .27	& .19	& .33	& .00 & .28 \\
& SVY\textsubscript{w} & .231 & .26 & .18 & .20 & .00 & .17 & .52 & .04 & .12 & .50 & .40 & .22 & .49 & .19 & .10 & .24 & .30 & .25 & .25 & .12 & .04 & .29 \\
\bottomrule
\end{tabular}
\caption{Achieved $F_1$-score of team Epicurus per test dataset (macro and for each of the 20 value categories). Our own macro $F_1$ is the unweighted average of the 20 individual $F_1$ scores, while the official macro $F_1$ is calculated using the averages of precision and recall over all 20 value categories. Approaches marked with * were not part of the official evaluation. Approaches in gray are shown for comparison: an ensemble using the best participant approach for each individual category, the best participant approach, and the organizer's BERT and 1-Baseline.}
\label{table-results}
\end{table*}

\subsection{Main Test Set}
On the main test set, all four models performed better than the two baselines. The best-performing model (ANN\textsubscript{w}) used annotations and weighted cross-entropy loss, achieving an $F_1$ score of .45 (15\% higher than the BERT baseline). The worst performing model (ANN\textsubscript{uw}) still achieved a 10\% increase in macro $F_1$ over the BERT baseline. Compared to the BERT baseline, the best performing model achieved substantially higher $F_1$ scores for stimulation (.13 vs .05), face (.29 vs .13) and humility (.21 vs .07), whereas prediction performance was worse for, amongst others, hedonism (.15 vs .20), security: personal (.70 vs .74), and conformity: interpersonal (.23 vs .19). Overall, on the main test set, the weighted models performed better than the unweighted models, and the models with annotation instructions performed better than those with survey questions. This is unsurprising, as using the annotation instructions probably more directly captures how the annotators came to label the data, whereas survey items are more distal. 

\subsection{Religious Texts Test Set}
Since the models were not trained on this data set, good prediction performance requires that the trained models generalize well to texts from a (very) different distribution. Three out of the four trained models performed better than the BERT baseline, and one model performed equally well. The best-performing model used survey questions and an unweighted loss function (SVY\textsubscript{uw}) and achieved an 18\% higher $F_1$ score than the BERT baseline. The pattern of model performance is different than on the main data set. Specifically, the models including survey items performed better than the ones including annotation instructions, which might indicate that using survey items (which are more distal measures of human values than annotation instructions) may help especially when predicting out-of-distribution arguments. The unweighted models performed better than the weighted ones, which is surprising. The best model achieved substantially higher $F_1$ scores compared to the BERT baseline for, amongst others, power: resources (.25 vs .00), face (.52 vs .28) and universalism: nature (.33 vs .00), and worse scores for universalism: tolerance (.00 vs .20) and hedonism (.40 vs .67).

Note that if we abide by the challenge organizers' macro $F_1$ calculation, the ranking of the models relative to each other and to the BERT baselines can be different. Especially on the test set comprising religious texts, per the organizers' calculation, none of our models outperformed the BERT baseline, and two models out of the four models achieved the same macro $F_1$ score as the BERT baseline.

\subsection{Influence of the Number of Definitional Statements on Macro $F_1$}
While our models achieved higher prediction performances than the challenge owners' baselines, a limitation of our approach is that, even with a relatively small number of training arguments/premises (<6,000), the total number of training instances can be very large, as this also depends on the number of value categories and definitional statements. In our experiments, one model took about 20 GPU hours to train. Computing times might become impractically long when the number of arguments, values and definitional statements increases.

Therefore, to investigate the scalability of our proposed approach, we explored the influence on the number of definitional statements per value category on our approach’s performance. Note that because these analyses were conducted after the challenge’s submission deadline, their results were not part of the official submissions. Additionally, in view of limited computational resources, we limited our additional analyses to the ANN\textsubscript{w} model. We tested ten different sample sizes – ranging from one to ten definitional statements per value category – with the respective number of definitional statements sampled using simple random sampling with replacements (to circumvent the issue of some value categories having a smaller number of definitional statements than the sample size of interest).

\begin{figure}[h]
\centering
\includegraphics[width=75mm,scale=0.7]{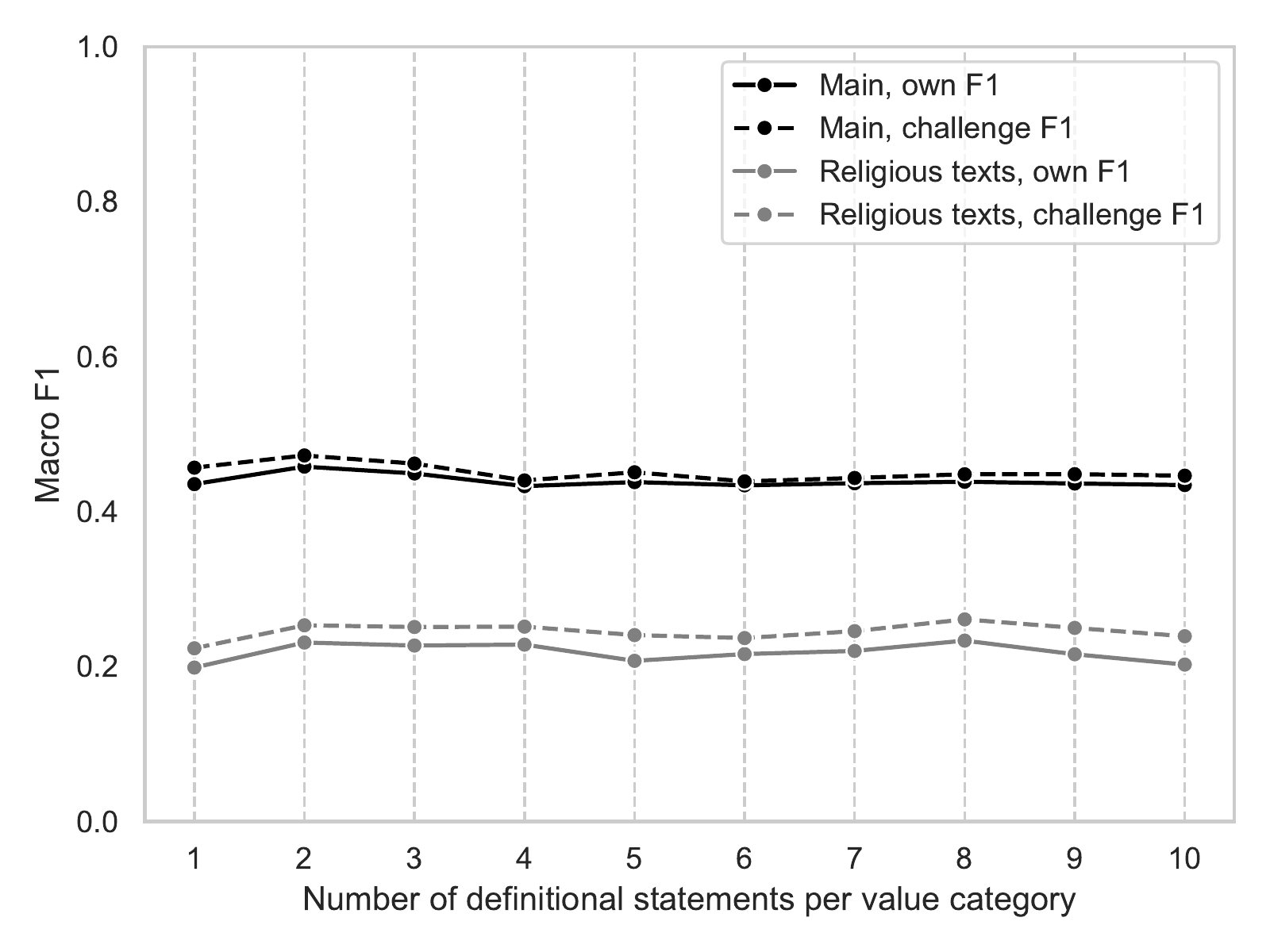}
\caption{Achieved macro $F_1$ on the main test set and religious texts test set, per number of definitional statements per value category.}
\label{figure4_additional}
\end{figure}

On the main test set, the highest macro $F_1$ was obtained with two definitional statements per value category (own macro $F_1$: 0.458; challenge organizer’s $F_1$: 0.472; see ~\ref{figure4_additional}). Macro $F_1$ decreased substantially as the number of definitional statements increased to four, and levelled off after that. We observed this trend for both macro $F_1$ calculation methods. On the test set comprising arguments from religious text, macro $F_1$ scores varied across the number of definitional statements. The best performance was obtained for a sample size of eight, while at sample sizes between two and four the achieved $F_1$ scores were already higher than the original ANN\textsubscript{w} model. 

These results show that even with just two or three definitional statements per value category, our proposed approach could achieve higher or comparable performance than when all available definitional statements are used, while the computational overhead is reduced by about 90 per cent. This suggests that our proposed approach is scalable by reducing the number of definitional statements per value category. 

For these additional analyses, we expected the models’ performance to increase with larger sample sizes, before eventually levelling off. However, the observed pattern was that, after peaking, performance decreased and then levelled off. A potential reason might be that most model hyperparameters (e.g., batch size, learning rate) were the same for all studied sample sizes (except for, for instance, step size, which depends on the training size), where these specific hyperparameter values might be potentially inappropriate for some of the models (especially those with substantially training sizes) to efficiently learn from the data.

\section{Conclusion and Discussion}
Our models achieved higher prediction performances than the challenge owners' baselines, indicating that there is merit to using definitional statements (i.e., annotation instructions and survey questions) for predicting the human values implied in textual arguments. This aligns with previous studies that incorporated dictionary definitions or survey questions for better task performance \citep[e.g.][]{kreuter2022items, fang2022modelling}. Furthermore, we showed that by using just a small number of definitional statements per value category, we could achieve prediction performance comparable to (and in some cases, better than) when all available definitional statements are used, while significantly reducing computational overhead.

As the goal of our study was not to obtain the best performance possible, but to test the idea that incorporating definitional statements into model training would improve prediction of human values, we did not try more advanced or larger language models, or techniques that could have improved prediction performance, such as paraphrasing \cite{wei2019eda} and ensemble learning. We also fine-tuned our models on only a limited set of hyperparameters.

A reason for why even the best team's model achieved a macro $F_1$ score of only .551 could be the low inter-rater agreement of the annotations. The average Krippendorff's $\alpha$ was just 0.49 for level-1 value categories (i.e., sub-values of level-2 categories) \cite{kiesel:2022}. To investigate the inter-rater agreement for the level-2 values, which are the focus of this challenge, two of the authors annotated a random sample of 100 arguments from the training data and arrived at an even lower $\alpha$ of just 0.31. Accordingly to Krippendorff \citeyearpar[p. 241]{krippendorff2004content}, $\alpha$ values below 0.667 reflect very poor inter-rater agreement and high random (measurement) error. A reason for this low inter-rater agreement might be the annotation scheme requiring annotators to classify arguments as relating to 54 values --- each of which with several instructions --- which can overwhelm even experienced annotators. Furthermore, some values and their associated explanations seem similar, such as "Power: resources" --- which partially concerns having wealth --- and "Security: personal" --- which partially concerns not having debts and having a comfortable life. Therefore, classifying an argument as belonging to a particular value may be more subjective than intended. Improvements in the human value taxonomy and/or the annotation scheme are likely needed to yield more reliable and valid measurements of human values.

\section*{Acknowledgement}
This work was partially supported by the Dutch Research Council (grant number VI.Vidi.195.152 to D. L. Oberski).

\bibliography{custom}
\bibliographystyle{acl_natbib}

\appendix

\section{Overview of Definitional Statements, Per Value Category}
\label{append_b}
See next page.

\begin{table*}[htb!]
\centering\small%
\setlength{\tabcolsep}{2.5pt}%
\begin{tabular}{@{}ll@{\hspace{10pt}}c@{\hspace{5pt}}cccccccccccccccccccccc@{}}
\toprule
\multicolumn{2}{@{}l}{} &  \rotatebox{90}{\bf Self-direction: thought} & \rotatebox{90}{\bf Self-direction: action} & \rotatebox{90}{\bf Stimulation} & \rotatebox{90}{\bf Hedonism} & \rotatebox{90}{\bf Achievement} & \rotatebox{90}{\bf Power: dominance} & \rotatebox{90}{\bf Power: resources} & \rotatebox{90}{\bf Face} & \rotatebox{90}{\bf Security: personal} & \rotatebox{90}{\bf Security: societal} & \rotatebox{90}{\bf Tradition} & \rotatebox{90}{\bf Conformity: rules} & \rotatebox{90}{\bf Conformity: interpersonal} & \rotatebox{90}{\bf Humility} & \rotatebox{90}{\bf Benevolence: caring} & \rotatebox{90}{\bf Benevolence: dependability} & \rotatebox{90}{\bf Universalism: concern} & \rotatebox{90}{\bf Universalism: nature} & \rotatebox{90}{\bf Universalism: tolerance} & \rotatebox{90}{\bf Universalism: objectivity} & \rotatebox{90}{\bf Total} \\
\midrule

Annotation instructions & & 18 & 17 & 15 & 6 & 26 & 11 & 7 & 9 & 28 & 12 & 12 & 13 & 8 & 12 & 28 & 11 & 18 & 18 & 12 & 13 & 294  \\

Survey questions & & 18  & 19 & 15  & 24  & 31  & 11  & 9  & 13  & 28  & 14 & 21 & 20 & 13  & 16  & 30  & 14  & 17 & 14  & 9  & 7 & 343 \\

Total & & 36 & 36 & 30 & 30 & 57 & 22 & 16 & 22 & 56 & 26 & 33 & 33 & 21 & 28 & 58 & 25 & 35 & 32 & 21 & 20 & 637 \\
\bottomrule
\end{tabular}
\caption{Overview of the number of annotation instructions, survey questions, and sum of the two, per value category.}
\label{table-definitionalstatements}
\end{table*}

\section{Other Implementation Details}
\label{append_a}

\textbf{Computing Infrastructure}
All analyses were done on one of the High Performance Computing (HPC) cluster offered by Utrecht University.
Python 3.10, PyTorch 1.12.1 \cite{pytorch}, torchtext 0.13.1,  Huggingface Transformers 4.24.0 \cite{wolf-etal-2020-transformers} and CUDA 11.3 were used for finetuning BERT \cite{devlin2018bert} and predicting the labels of the test set. Numpy 1.13.1 \cite{harris2020array}, pandas 1.4.4 \cite{reback2022pandas}, and scikit\_learn 1.1.3 \cite{scikit-learn} were used for data wrangling.

\textbf{Runtime}
About 20 hours per model on an RTX 6000 GPU node. 

\textbf{Number of parameters}
110 million.

\textbf{Validation performance}
ANN\textsubscript{uw}: 0.432; ANN\textsubscript{w}: 0.441; SVY\textsubscript{uw}: 0.423; SVY\textsubscript{w}: 0.434.

\textbf{Hyperparamters}
For the BERT models, we used the following hyperparameters:

- num\_train\_epochs=5

- per\_device\_train\_batch\_size=128

- per\_device\_eval\_batch\_size=1024

- warmup\_steps=250

- weight\_decay=0.01

- early stopping criterion: step

- patience for early stopping: 10

\end{document}